\documentclass[a4paper,twoside]{article}

\usepackage{epsfig}
\usepackage{subcaption}
\usepackage{calc}
\usepackage{amssymb}
\usepackage{amstext}
\usepackage{amsmath}
\usepackage{amsthm}
\usepackage{multicol}
\usepackage{xcolor}
\usepackage{hyperref}
\usepackage{caption}
\usepackage{pslatex}
\usepackage{apalike}
\usepackage{textcomp} 
\usepackage{graphicx}
\usepackage{array}
\newcolumntype{P}[1]{>{\centering\arraybackslash}p{#1}}
\newcolumntype{M}[1]{>{\centering\arraybackslash}m{#1}}
\usepackage{comment}

\usepackage{SCITEPRESS}     % Please add other packages that you may need BEFORE the SCITEPRESS.sty package.

\begin{document}

\title{Assessing COVID-19 Impacts on College Students via Automated Processing of Free-form Text}

\author{
\authorname{Ravi Sharma\sup{1},  
Sri Divya Pagadala\sup{2},  
Pratool Bharti\sup{2},
Sriram Chellappan\sup{1},
Trine Schmidt\sup{3} and 
Raj Goyal\sup{3}}
%\orcidAuthor{}
\affiliation{\sup{1}Department Of Computer Science and Engineering, University Of South Florida, Tampa, FL, USA}
\affiliation{\sup{2}Department of Computer Science, Northern Illinois University, DeKalb, IL, USA}
\affiliation{\sup{3}Ajivar LLC, Tarpon Springs, FL, USA}
\email{\{ravis, sriramc\}@usf.edu, \{spagadala1, pbharti\}@niu.edu, \{trine, raj\}@ajivar.com}
}


\abstract{In this paper, we report experimental results on assessing the impact of COVID-19 on college students by processing free-form texts generated by them. By free-form texts, we mean textual entries posted by college students (enrolled in a four year US college) via an app specifically designed to assess and improve their mental health. Using a dataset comprising of more than $9000$ textual entries from $1451$ students collected over four months (split between pre and post COVID-19), and established NLP techniques,  a) we assess how topics of most interest to student change between pre and post COVID-19, and b) we assess the sentiments that students exhibit in each topic between pre and post COVID-19. Our analysis reveals that topics like Education became noticeably less important to students post COVID-19, while Health became much more trending. We also found that across all topics, negative sentiment among students post COVID-19 was much higher compared to pre-COVID-19. We expect our study to have an impact on policy-makers in higher education across several spectra, including college administrators, teachers, parents, and mental health counselors.}

\onecolumn \maketitle \normalsize \setcounter{footnote}{0} \vfill

\section{\uppercase{Introduction}}
\label{sec:introduction}

\noindent The COVID-19 pandemic has impacted people across the globe irrespective of where they live, how old they are, and what they do. From a physical health perspective, COVID-19 has affected senior citizens the most. However, from a mental health perspective, younger people (especially college students) have been severely impacted due to many factors including transitioning to new modes of instruction, loss of friend circles, cancellation of classes and even entire semesters,  financial hardships, shrinking job markets, failing relationships and much more. Also, since mental health services are harder to access now (due to social distancing necessities), students are suffering even more from this pandemic.

On a related note, various stakeholders in higher education are also impacted because there is a need to manage new forms of online instruction, understand and react to the financial hardships of students, ensure classrooms are socially isolated, generate facilities for  COVID-19 testing, and more. In overcoming these challenges, there needs to be a mechanism to understand student needs and expectations in real-time, as students are the most important, while also being vulnerable entities in the higher education ecosystem. 

One approach to get student perceptions in such times is to send periodic surveys to them. But these can be sent only a limited number of times to prevent students from simply choosing not to respond. Survey responses also suffer from well-known recency biases \cite{recency}. Furthermore, surveys cannot proactively capture student perceptions and are sensitive more to the mindset of those creating the surveys, rather than students themselves. 

The next viable option to glean student perceptions is mining social media. This is an area rich in exploration today, and there are many  studies on mining social media content to understand student mental health needs \cite{mental1} \cite{mental2}, relationship tendencies \cite{relation}, academic performance \cite{academic}, political participation \cite{politics1}, \cite{politics2} and much more. But again, the core issue with mining social media data in the context of our problem in this paper stems from a) social media data being noisy, meaning that there is so much irrelevant information out there, which makes it much harder to filter out and extract only ones relevant to any particular scenario; b) the fact that there are so many social media sites that are popular today, and so it becomes increasingly harder to make conclusions by processing data from only one or few of them. Additionally, these sites are very hard for developers to crawl massive data due to privacy concerns.

In this paper, we report the results of an experiment we conducted at a four-year college in the US to assess topics and sentiments of students before the COVID-19 pandemic began and after, via processing free-form texts collected via a smartphone app specifically designed for students to express their moods, feelings, opinions via the app. %The app has been in use by students since Aug $2019$. For this study, we collected more than $9,000$ posts from  $500$ students starting from February $2020$ to April $2020$. This period is most relevant to this study since the COVID-19 pandemic hit the US during the middle of March $2020$. 
With this dataset, we wanted to answer, broadly speaking, two questions presented below:
%are reactive in the sense that and one that is used, its limitations are patent - including lack of interest from students upon discussions with many mental health counselors in various US campuses, we found out that counselors are most interested in knowing what are the sources that 
\begin{enumerate}
    \item {Before COVID-19, what are the topics that students most posted about, and how did the topics change post-COVID-19?}
    \item{Across these topics and timelines, how did the sentiment of students change between positive, neutral, and negative?}
\end{enumerate}
%subsection{Structure of the paper}
To answer the above questions, we process our dataset using several standard and well-established techniques in Natural Language Processing. Our analysis that compares the post-COVID-19 timeframe with the pre-COVID-19 timeframe revealed Health became a highly trending topic for our subjects which increased sharply post COVID-19 timeframe while Housing, Finance, and Relationships experiencing a marginal increase. Also, Education decreased significantly during the post-COVID-19 timeframe. From the perspective of emotion, universally across each topic, there was a significant increase in negative sentiment in the post-COVID-19 timeframe.

To the best of our knowledge, this study is unique in processing free-form text generated via a custom made app (that was designed to glean mental health insights of college students) to infer trending topics and sentiments in the COVID-19 timeframe. Towards the end of the paper, we present how our work enables policy-makers in higher education to cater to student mental health and other needs in these unique times. Additionally, the data will also help the app developers to optimize the AI to deliver individualized resources that help the specific student in their time of need based on their sentiment, etc. 

The rest of the paper is organized as follows. Section $2$ presents related work in the literature. Section $3$ explains our dataset. Section $4$ details our underlying research methodology. Section $5$ presents our results. Finally, Section $6$ presents a related discussion, conclusions, and future work.
\vspace{-1em}
\section{\uppercase{Related Work}}
\label{sec:relatedWork}
\noindent Since December 2019, when the COVID-19 pandemic began, there has been an urgent interest across the globe to understand its impact from many perspectives. In particular, the genetic aspects of the disease, the epidemiological aspects, the impact of contact tracing, the impact on economies, and political aspects across the globe are all actively being investigated today.  In this section though, we highlight current research in the realm of COVID-19 that is most relevant to our work in this paper, namely the processing of texts over the Internet concerning the pandemic.

In \cite{jelodar}, researchers applied topic modeling on COVID-19–related Reddit comments to extract meaningful categories from public opinions. They then designed an LSTM recurrent neural network \cite{lstm} for polarity and sentiment classification. They used $sentiStrength$ \cite{thelwall2} for sentiment detection and average out the sentiment of COVID-19 related comments obtained from 10 subreddits forums. They labeled the data as per $sentiStrength$ scores and achieved an $81.15\%$ classification accuracy. In \cite{oyebode}, comment data obtained from social media is processed and then context-aware themes are extracted. For each theme, the sentiment is extracted using VADER lexicon-based algorithm \cite{vader} and only themes having positive or negative scores are kept, neutral themes are discarded.  Later, these themes are categorized into 34 negative and 20 positive broader themes using human reviewers. These themes span different areas related to health, psychological, and social issues and showed the impact of COVID-19 negatively. The authors also recommend possible solutions for negative themes. In another paper \cite{kabir2020coronavis}, the authors also mine Twitter to use collected tweets as a repository to analyze information shared by the people to visualize topics of interest, and the emotions of common citizens in the USA as the pandemic is unfolding.

Our work in this paper broadly falls in the above categories. However, there are compelling differences. First, in existing research on processing social media data, the trend is to ignore the demographics of subjects, since demographic information is very hard to find from social media data, and even if a person does post demographic information, there is little evidence to check if the posted information is correct. So, if the goal is to understand the topic and emotions of a particular demographic only, then mining social media data can be problematic. In this paper, our goal is to {\em only} look at college students as they face the pandemic, and hence we did not want to mine social media data. Instead, we mine free-form texts that students post using an app, that is {\em specifically} designed to monitor and enhance student mental health. As such, the reliability of our data, as it pertains to our demographic is significantly high. Secondly, with social media data, there is a massive scale of irrelevant information, and during pre-processing, there is always a danger of discarding contextually relevant information and/or incorporating contextually irrelevant information. Both of these things are avoided in our data since again, the context of the app is promoting the mental health of students. Hence we believe that our dataset and corresponding results in this paper, and unique and contextually very relevant, in terms of gleaning insights on how college students (in the USA) are responding to the COVID-19 pandemic.
%\textcolor{blue}{Wrap up a paragraph: Need Your suggestion on this} Unfortunately, though, there is significant clutter in social media logs, and the risk of processing irrelevant posts, posts tagged with incorrect location information, fake news, etc. is a major challenge. In our study, we followed a more directed and task-specific approach and collect and analyze college students' free-texts about their issues. This, not only significantly minimized the computing overhead while pre-processing data and removing the noise as compared to social media logs, but also gives the students free hand to express their perception in a focused university setting. In the next section, we discuss our data collection and cleaning process. 
\vspace{-2em}
\section{DATA DESCRIPTION AND CLEANING}
%We believe the free text data collected is novel. The data provided can range to not only different domains but also show varying emotions as well. Also, the first study was conducted on US students impacted by COVID-19. [TODO: LOTS OF REFERENCES, Change this entirely]
\subsection{Data Description}
Data (i.e., free-form text) for this study was collected in partnership with Ajivar LLC. \cite{ajivar}, a company that designs smartphone technologies to assess and enhance the Emotional Intelligence (EI) of students, so that they are more resilient to life's challenges. At the core of Ajivar's capabilities is an engaging smartphone app that students can download, which enables them to compose {\em free-form} texts in English to the app, either on their own or in response to a request via the app. During interaction with Ajivar, students share anything from their day to day life experiences, problems, happy/sad moments, situations revolving around their academic and personal lives, and so on. These texts are then processed by Ajivar’s AI, in an automated manner, so that real-time empathetic responses are given to students to improve their EI. This service while being important at any time to college students becomes even more important during this pandemic time. As of today, more than $9000$ students, at a four-year US college are using Ajivar's app that is available on both iOS and Android.

\begin{figure}[!htbp]
    \centering
    \includegraphics[width=\linewidth]{"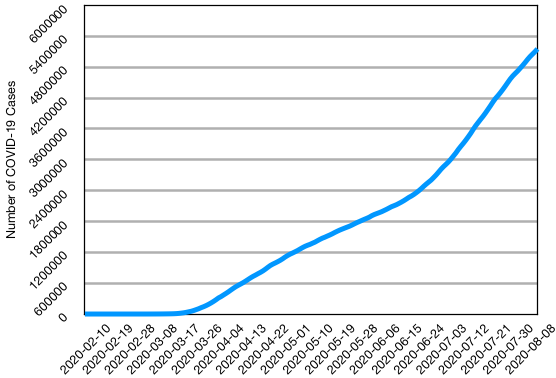"}
    \caption{COVID-19 spread in US in $2020$}
    \label{fig:cdc}
\end{figure}

For the purposes of this study, $9090$ free-form texts (all in English) were collected from $1451$ students between February $1$, $2020$ and April $30$, $2020$. This timeframe is ideal for this study because March $15$ was about the time, the COVID-19 pandemic was considered serious enough in the USA (as seen in Fig. \ref{fig:cdc}), that necessitated significant changes in the lives of many in the US (and especially college students). We had an equal number of texts before and after March $15$ in our dataset to keep it balanced pre and post COVID-19. Some examples of texts in our dataset are presented in Table  \ref{tab:dataCleanPost} that we used in this study. All personal identifiers were removed manually before processing the texts.
\vspace{-1em}
\subsection{Data Cleaning}
\label{subsec:dataCleaning}
\noindent
Due to the free-form nature of data, it can be noisy and requires cleansing before further processing. To do so, we processed texts in our dataset (both pre and post-COVID-19) to remove sources of noise including stopwords and generic terms using the popular Natural Language Toolkit (NLTK) \cite{nltk}. Due to the nature of free-form texts, we expected some un-intelligible words in our dataset like `gjhgdjfg', `jshdfkj', etc. However, we did not encounter any such words. Typos were corrected using publicly available spellchecker software. 

We then performed lemmatization with part of speech (POS) tags to convert the remaining (and relevant) words to their root form for standardization during processing. For instance, Lemmatization step will convert ``booking" to book (since booking is a verb) but will convert ``string" to ``string" again (since string is a noun). 
\begin{figure}[h]
    \includegraphics[width=\columnwidth]{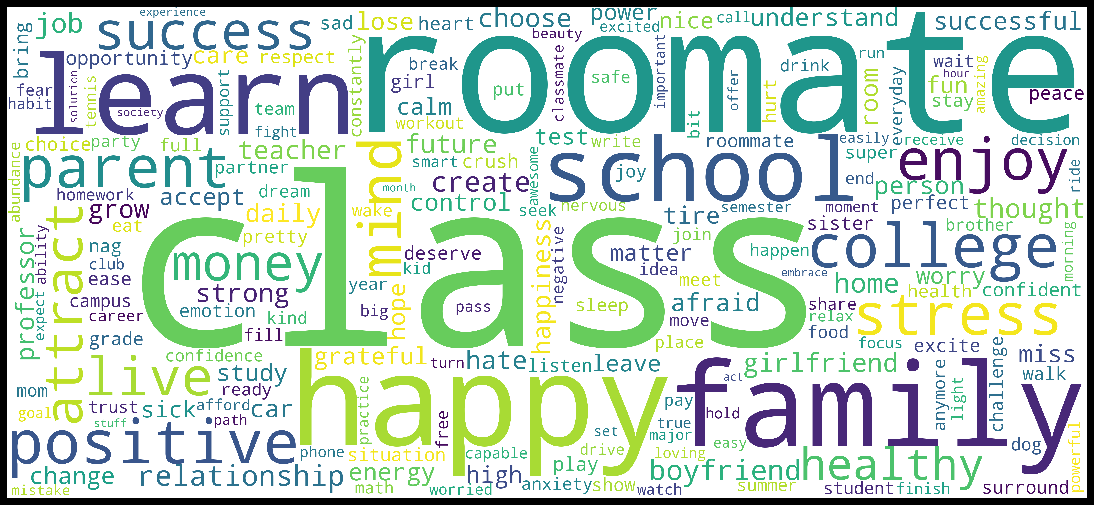}
    \caption{Word-cloud for pre-COVID-19 Data}
    \label{fig:pre_wc}
\end{figure}
\begin{figure}[!h]
    \centering
    \includegraphics[width=\columnwidth]{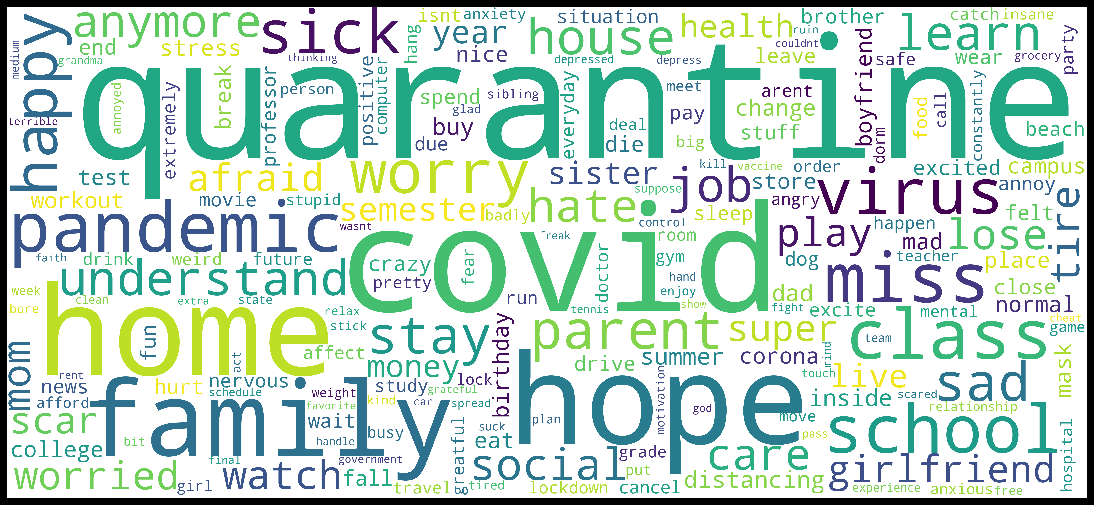}
    \caption{Word-Cloud for post-COVID-19 Data}
    \label{fig:post_wc}
\end{figure}

Finally, for illustration purposes, table \ref{tab:dataCleanPost} shows the result of the data cleaning process. In Fig. \ref{fig:pre_wc} and Fig. \ref{fig:post_wc}, we show word clouds for the cleaned data in both pre and post COVID-19. We can clearly see how words like ``class", ``roommate", ``learn", ``enjoy" and ``family" were dominant in the pre-COVID-19 phase, compared to words like ``quarantine", ``covid", ``sick", ``hope" and ``pandemic" that were dominant post-COVID-19. Presenting these results in a more formalized context from the perspective of identifying the most critical topics for students and their associated sentiments as a result of the pandemic is our end goal and is presented next in detail. 
\vspace{-1em}
\begin{table*}[!h]
    \centering
    \caption{Data Cleaning Process}
    %\caption{Data Cleaning Process for Post-COVID Data}
    \label{tab:dataCleanPost}
    \begin{tabular}{|M{0.65\columnwidth}|M{0.63\columnwidth}|M{0.63\columnwidth}|}
         \hline
         \textbf{Students' Texts} & \textbf{Removal of Stopwords and Generic Terms} & \textbf{Final Text After Lemmatization} \\  \hline
            Freaking out my grades ever since online learning. &	[`Freaking', `grades', `online', `learning'] &	[`freak', `grade', `online', `learning']	\\  
            \iffalse
            \hline	
            I am tired of online classes &	[`tired', `online', `classes'] &	[`tire', `online', `class']	\\  \hline
            I've had so many unnecessary food expenses because of quarantine &	[`unnecessary', `food', `expenses', `quarantine'] &	[`unnecessary', `food', `expense', `quarantine']	\\  \hline
            My doctor just got covid and I was at his office this week I am very scared I will get it too  &	[`doctor', `covid', `office', `week', `scared'] &	[`doctor', `covid', `office', `week', `scare']	\\  \hline
            I am tired of being home with my family because my parents and siblings are annoying  &	[`tired', `home', `family', `because', `parents', `siblings', `annoying'] &	[`tire', `home', `family', `because', `parent', `sibling', `annoy']	\\  \hline
            My girlfriend just broke up with me because my mom wouldn't let me leave the house so she found another guy who could hang out with her  &	[`girlfriend', `broke', `mom', `leave', `house', `found', `guy', `hang'] &	[`girlfriend', `break', `mom', `leave', `house', `find', `guy', `hang']	\\  
            \fi
            \hline
            Im constantly thinking about domestic abuse victims who have to stay in their homes now  &	[`constantly', `thinking', `domestic', `abuse', `victims', `stay', `homes'] &	[`constantly', `think', `domestic', `abuse', `victim', `stay', `home']	\\  \hline
            %I am worried that my boyfriend will fall out of love with me because I am still not allowed to see him  &	[`worried', `boyfriend', `fall', `allowed'] &	[`worry', `boyfriend', `fall', `allow']	\\  \hline
%            I lost my housing due the virus  &	[`lost', `housing', `due', `virus'] &	[`lose', `housing', `due', `virus']	\\  \hline
            My sleep schedule is off track because im inside all day.  &	[`sleep', `schedule', `track', `inside'] &	[`sleep', `schedule', `track', `inside']	\\  \hline
    \end{tabular}
\end{table*}
%\vspace{-1em}
\begin{table*}[!ht]
\caption{Top $20$ Frequent Words for Pre and Post COVID-19}
    \label{tab:wordFreq}
    \centering
    \begin{tabular}{|M{0.2\columnwidth}|M{0.21\columnwidth}|M{0.21\columnwidth}|M{0.2\columnwidth}|M{0.2\columnwidth}|M{0.2\columnwidth}|M{0.2\columnwidth}|M{0.2\columnwidth}|}
 \hline
 \multicolumn{4}{|c|}{\bf{Pre-COVID-19}} &
  \multicolumn{4}{c|}{\bf{Post-COVID-19}} \\
 \hline
        {Word} & {Frequency} &
        {Word} & {Frequency} &
        {Word} & {Frequency} &
        {Word} & {Frequency} \\
        \hline
        class & 90 & roomate & 71 & quarantine & 277 & covid & 210  \\ 
        happy & 70 & family & 69 & hope & 147 & home & 145  \\  
        learn & 62 & school & 56 & family & 145 & pandemic & 128 \\
        positive & 53 & college & 52 & miss & 126 & class & 125 \\
        mind & 50 & stress & 49 & school & 116 & virus & 105 \\ 
        parent & 49 & enjoy & 48 & happy & 100 & worry & 99 \\ 
        live & 41 & attract & 41 & sick & 99 & parent & 98 \\ 
        success & 40 & money & 40 & sad & 96 & stay & 92 \\
        relationship & 39 & healthy & 39 & house & 84 & learn & 81 \\    
        create & 37 & boyfriend & 37 & understand & 80 & job & 78 \\ 
        \hline
        
    \end{tabular}
\end{table*}

\begin{table*}[!ht]
    \caption{Topics and Vocabulary for categorization and Semantic Similarity}
    \label{tab:category}
    \centering
    %\begin{tabular}{|c|c|}
    \begin{tabular}{|M{0.25\columnwidth}|M{1.72\columnwidth}|}
         \hline
         \textbf{Topics} & \textbf{Vocabulary}\\  
            \hline
            Education & grade, lecture, professor, exam, attendance, homework, quiz, assignment, syllabus, gpa, marks, college, school, study,  \\  
            \hline
            Finance & tuition, job, money, rent, debt, grocery, fees, wage, dollars, shopping, car, budget, scholarship, cash, salary, bills, economy, poor, rich, income \\
            \hline
            Health & health, disease, illness, nutrition, vaccine, medicine, diagnosis, sickness, eat, food, disability,  hospital, exercise, gym, covid, pandemic, quarantine, lockdown, virus, corona \\  
            \hline
            Family & mom, dad, grandma, brother, sister, grandparents, aunt, uncle, sibling, son, daughter, kid, wife, relative, family, cousin, stepfather, stepmother, husband, home, homesick, fiance \\
            \hline
            Relationships & boyfriend, girlfriend, bf, gf, crush, friend, buddy, dude, ex, stranger, people, partner, associate, pal, mate  \\ 
            \hline 
            Housing & hostel, apartment, roommate, house, dorm, cloths, room, bed, window, bathroom, fan, sound \\ 
            \hline
    \end{tabular}
\end{table*}
%the cleaning process for pre and post COVID-19 data. Post data cleaning, we obtained the texts which were free from stopwords and generic terms. Further, they were transformed into their root words. We used these root words to create a word cloud that reflects the importance of each word in the given context. A word cloud is a visualization technique to display a set of words in which the size of each word indicates its frequency or importance in the dataset. We created a word cloud based on pre and post COVID-19 root words to observe the difference in important terms in each category. Fig. \ref{fig:pre_wc} and fig. \ref{fig:post_wc} shows the word-cloud for most frequent words for pre and post COVID-19 data respectively after applying cleaning steps. As we can see, words occurring in the post-COVID-19 setting are more related to pandemic (i.e. quarantine, covid, virus, hope, etc.), while words in the pre-COVID-19 cloud are more aligned with general academics (i.e. class, roommate, learn, school, etc.). %% \subsection{TOP WORDS BY FREQUENCY}

\section{OUR METHODOLOGY FOR TOPIC CATEGORIZATION AND SENTIMENT ANALYSIS}
\noindent In this section, we present the overall methodology for automating the topic categorization and sentiment determination from the perspective of COVID-19 and college students. Recall the two major questions we want to answer in this study: 1) Before COVID-19, what are the topics that students most posted about, and how did these topics change post-COVID-19? 2) Across these topics, how did the sentiment of students change between positive, neutral, and negative?

\iffalse
The main goal of this study is to analyze the impact of COVID-19 on college students in various contextual categories (i.e., education, finance, health, family, etc.). By leveraging the text analysis and data mining techniques on students' data, we want to seek answers to the following questions:
\begin{enumerate}
    \item What are the major categories (topics) prevalent in the students’ academic life?
    \item How is COVID-19 affecting college students in each of these prevalent categories?
    \item How is COVID-19 affecting students' sentiments within these categories?
\end{enumerate}
\fi

\subsection{Determining Topics of Interest Via Contextual and Semantic Categorization}
To answer the first question, we perform two critical steps - Contextual Categorization and Semantic Categorization. Step $1$ is to use our intuition, domain expertise of counselors at Ajivar, and the faculty/student authors of this paper to categorize the broad topics of interest to college students as it pertains to this pandemic. However, instead of merely relying on our intuition, we also delved deeper into the output of the Data Cleanup process (presented earlier), namely the root words of students' free-form texts. Table \ref{tab:wordFreq} presents the top $20$ root words in both the pre and post COVID-19 phases after lemmatization. Combining our intuitions, categories mentioned in related works \cite{category1} \cite{category2}\cite{category3}\cite{alison} and coupled with the results gleaned from student posts in Table \ref{tab:wordFreq}, we identified a total of six topics along with their corresponding vocabulary presented in Table \ref{tab:category} that we believe are most relevant to college students. This completes Step $1$ - the identification of topics of interest to students.

The next step towards answering Question $1$ involves building an AI model to {\em semantically} map a student's post to a topic. To do this, we need to capture the semantic textual similarity between the students' texts and topics. Transformers \cite{transformers} based models have been a popular choice for performing various NLP tasks including semantic textual similarity in recent years due to their improved performance over existing methods. 

This had lead to the development of pre-trained systems like Bidirectional Encoder Representations from Transformers (BERT) \cite{bert}, where pre-trained representations capture a large spectrum of contexts and generate embeddings of each word based on the underlying context in the sentence. RoBERTa  \cite{roberta} is an improvement to the BERT model using larger training data with longer sequences, and dynamic masking. It has achieved achieve state-of-the-art performance on Multi-Genre Natural Language Inference Corpus (MNLI) \cite{mnli}, Question-answering NLI (QNLI) \cite{glue}, Semantic Textual Similarity Benchmark (STS-B) \cite{stsb}, ReAding Comprehension dataset from Examination (RACE) \cite{race} and GLUE \cite{glue}. 

Though both BERT and RoBERTa are recent and state of the art, the sentences to be checked for similarity before feeding into the network resulting in computational overhead. This can lead to scaling issues later on. To leverage state-of-the-art performance by BERT/RoBERTa but avoiding computational overhead, we leverage an extremely robust and more recent neural network architecture called Sentence-BERT (SBERT) \cite{sen-bert}. \\
\indent The core idea of SBERT is the notion of Sentence-level Embedding via pre-trained Transformers-based models (e.g. BERT and RoBERTa). The notion of Sentence-level embedding is to embed sentences into vector representations that retain the context of each word within the sentence (instead of the actual word), and have been a popular choice for semantic textual similarity over recent years. For instance, robust, semantic embeddings for two sentences like ``We are very happy now." and ``Our group is enjoying presently." will be very similar, even though they contain a completely different set of words. Needless to say, such, robust embedding is critical for our problem, since two free-form texts from students may have words completely different from each other, but still could refer to the same topic, for example, sentences like ``Have a homework to submit tonight." and ``Working on an assignment due later today", despite having no common words must have similar semantic embeddings since both refer to the topic of Education. How to do this effectively is our challenge. \\ \indent For this challenge, we use SBERT that improves upon BERT and RoBERTa, by optimizing semantic textual similarity using sentence-level embedding from  siamese and triplet network structure \cite{schroff}. This allows the model to derive fixed-size vectors for input sentences, and are much more efficient to scale. In our study, posts corresponding to pre and post COVID-19, along with the vocabulary for each Topic identified in Table \ref{tab:category} was fed into the SBERT network to derive sentence-embeddings. 
Once we derive semantic vector embeddings, we compare pre and post text embeddings with embeddings generated from words for each Topic in Table \ref{tab:category} using the notion of Cosine similarity \cite{cosine}. The output of this step essentially gives a notion of similarity between the two embeddings compared. For every text, the topic which it belongs to was identified as the one for whose embedding, the similarity with the embedding of the corresponding text was identified as the highest. 
\subsection{Sentiment Analysis of Topics Pre and Post COVID-19}
\noindent
We now are ready to detail our method to glean the sentiment of a post both pre and post COVID-19. For a comprehensive analysis, multiple states of art NLP models were used for validity analysis,  namely, VADER  and TextBlob: %and Transformers based models:  
\subsubsection{VADER}
\noindent Valence Aware Dictionary and sEntiment Reasoner (VADER) \cite{vader} is a lexicon and rule-based sentiment analysis tool.  It does not require any training data and is constructed from a generalized, valence-based human-curated gold standard sentiment. It is fast and gives positive, negative, neutral and an overall sentiment score. It is especially effective for more casual posts (as in free-form text) and is especially useful for mining sentiments of social media posts, YouTube comments, etc.

\subsubsection{TextBlob} TextBlob \cite{textblob} is a Python-based open-source library and returns polarity and subjectivity of a text input. The polarity is in the float range $[-1,+1]$ with -1 being highly negative and $+1$ being highly positive sentiment. Subjectivity lies in the range $[0,1]$ where $0$ being objective and $1$ being subjective. The way it works is that TextBlob goes along finding words and phrases it can assign polarity and subjectivity to based on predefined rules, and it averages them all together for longer text. It is especially useful for more formal texts.
%\subsubsection{Transformers-HuggingFace} 
%\noindent Transformers based methods give a state-of-the-art performance on various NLP tasks such as text classification, summarization, etc. Several pre-trained models exist which utilize the transformers architecture. We used the Transformers library by HuggingFace \cite{huggingFace} and leverage the Roberta-For-Sequence-Classification for sentiment analysis using  \cite{vicd}. This returns positive and negative sentiment values which are pre-trained with 768 hidden layers, a vocabulary of size 50k lexicons, and positional embeddings of size 514. %The motivation behind using the transformer-based method is because of the shortage of task-specific datasets. 
\vspace{-1em}
\section{\uppercase{Results}}
\label{sec:results}
\noindent In this section, we discuss the results of topic distribution identification and sentiment analysis in the pre vs post-COVID-19 dataset.
\subsection{Topic Categorization and Distribution}
\begin{figure}[htbp]
    \centering
    \includegraphics[width=\columnwidth]{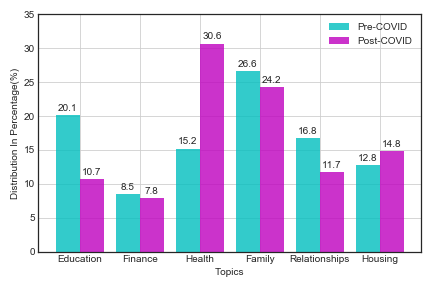}
    \caption{Topic Distribution  Comparison Between Pre and Post COVID-19}
    \label{fig:distribution}
\end{figure}

Fig. \ref{fig:distribution} presents our results of topics of importance to students before and after COVID-19. We see from the figure that there is a significant shift in the importance of these topics before and after COVID-19. We see that ``Education" had a steep drop as a topic in students’ posts after COVID-19 compared to before COVID-19. After COVID-19, it had the second-lowest importance (after Finance) among students. The ``Health" topic has the biggest jump among student concerns in post-COVID-19 timeframe compared to the pre-COVID-19 timeframe. There was not much change in the ``Family" topic, but we do see an increase in ``Housing" and a decrease in ``Relationship" topics in the post-COVID-19 timeframe compared to the pre-COVID-19 timeframe.
\begin{figure*}[!h]
    \centering
    \includegraphics[width=\textwidth,height=8cm]{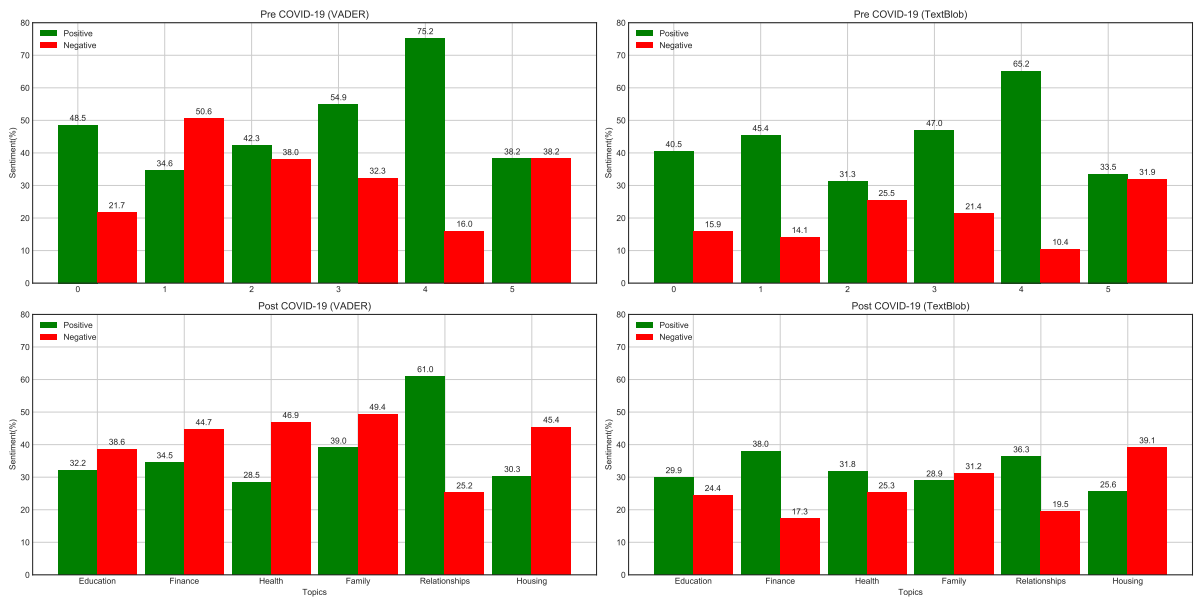}
    \caption{Sentiment across Topics Pre and Post COVID-19}
    \label{fig:sentiments}
\end{figure*}
\vspace{-1em}
\subsection{Sentiment Analysis among Topics}
While the topics of interest to students are important for decision-makers in academic environments, it is important to also glean what is the sentiment expressed by students on these topics, and how they change over time. For each of the two sentiment analysis models we utilized, the corresponding sentiments for the topics are presented in Fig. \ref{fig:sentiments}. The results from left to right are for VADER and TextBlob. At the top, is the pre-COVID-19 timeframe, and at the bottom is the post-COVID-19 timeframe. In general, overall sentiment values saw a decrease in positive sentiment and an increase in negative sentiment. 

\vspace{-1em}
\section{\uppercase{Discussion \& Conclusions}}
\noindent COVID-19 is still adversely affecting the lives of college students, in the form of numerous uncertainties from the perspective of health, finance, career, relationships, and more, that has taken a toll on their mental health. It is important to understand their issues at the micro and macro levels to provide an effective and scalable solution.
% What is important though is to understand student needs both at the micro and the macro level. Because, unless decision-makers have the correct insights on the problem, solutions will be unplanned and not accommodate to student needs, especially when resources are scarce, and there is no end to the problem in sight. 
Our paper makes important early contributions to this space in terms of gleaning insights on student needs from a custom app, designed and deployed to detect and address student mental health concerns. For instance, our finding that ``Education" is not a topic of critical interest to students posts COVID-19 should raise concerns on administrators to ensure that students are motivated to learn and that the quality of instruction must be given priority. Our findings related to how topics of interest changed post-pandemic are also important for administrators to know, to channel their resources in the appropriate directions.

Also, the majority of declination of categories from pre to post-covid-19 such as ``Education'', ``Relationships'', and ``Family'' is due to the sharp increase in the ``Health'' category which further emphasis the fact that medical concerns and other related topics were affecting the conversations of the students. 
%In this paper, we have gleaned the information about college student's primary areas of concern during COVID-19. To do so, we analyzed the real data collected from college students during the pre and post COVID-19 period and exhibited how their topics of discussion have changed and how much. We observed that, while education was one of the tops discussed topics in pre-COVID posts it got overshadowed by health, relationships, housing, and family discussion. Additionally, we also showed that negative sentiments are dominant in post-COVID-19 posts. Our findings can help the university and government personnel to identify the major pain areas for students and allocate resources in an appropriate direction in real-time. Since it is clear from our results that students are not highly concerned about their education, faculties can use our findings to address some of the prominent concerns in their classroom that can help students to re-focus on studies and keep them positive.
Although our findings are critical to understanding the college students’ areas of concern, it is important to mention the limitations and underline assumptions of our study. 

Firstly, all the participants in the study belong to the same university and hence can have similarities in their context to some degree. This also limits the study to issues related to geographical demographics of one type. Secondly, the free-texts provided by the participant were not categorized by the age groups and were merged, which may overlook the issues specific to graduate vs undergraduate students. Next, the population considered in our study were overlapped for the two-time frames meaning pre and post-COVID-19 data can be from the same or different students.  This restricts us from conducting any statistical studies on the samples as the style and structure of the texts can vary from person to person and for a generalized distribution for categories and sentiments, clear separation between users is needed. We would like to mitigate these issues by collecting more data across universities and demographics.
% Finally, the texts provided by students are free and thus can overlap between two or more categories. 

% In our future work, we would like to address the above-mentioned issues. First, our goal is to collect more data across the university. This includes the data with various settings like comparing the same student population data for pre and post-Covid-19 and comparing different populations across the time settings. For pre-COVID-19 data, we will be referencing legacy data sources already present in the literature for students. We want to conduct a detailed comparison of graduate vs undergraduate students for the experiments. 
In conclusion, authorities in universities have outreach programs, counselors, and other traditional measures to ease out the mental distress among students. While these are adaptable for the academic persona in general, it can be ineffective with scaling student population, lack of counselors, hesitant target subjects, etc. An effective measure to merge with the traditional methods is to curb these problems and automate the process.  Methods to provide students with comfortable sharing and categorize them in defined classes for easy management can not only solve scaling issues of growing student mental health cases but also tackles the lack of counselors’ problem. 
Though our study is not exhaustive to all students and should not be taken as a systematic assessment, we believe that can be a preliminary step towards providing insight to policymakers, stakeholders, and administrative figures to develop strategies for students' needs during the pandemic. 

\bibliographystyle{apalike}
{\small
\bibliography{main}}

\begin{thebibliography}{}

\bibitem[Ajivar, 2019]{ajivar}
Ajivar (2019).
\newblock Life coach powered by artificial intelligence, https://ajivar.com.

\bibitem[AlKandari, 2020]{category1}
AlKandari, N.~Y. (2020).
\newblock Students anxiety experiences in higher education institutions.
\newblock In {\em Anxiety Disorders}. IntechOpen.

\bibitem[Bagroy et~al., 2017]{mental1}
Bagroy, S., Kumaraguru, P., and De~Choudhury, M. (2017).
\newblock A social media based index of mental well-being in college campuses.
\newblock In {\em Proceedings of the 2017 CHI Conference on Human factors in
  Computing Systems}, pages 1634--1646.

\bibitem[Brooker et~al., 2017]{category3}
Brooker, A., Brooker, S., and Lawrence, J. (2017).
\newblock First year students' perceptions of their difficulties.
\newblock {\em Student Success}, 8(1):49--63.

\bibitem[Cer et~al., 2017]{stsb}
Cer, D., Diab, M., Agirre, E., Lopez-Gazpio, I., and Specia, L. (2017).
\newblock Semeval-2017 task 1: Semantic textual similarity-multilingual and
  cross-lingual focused evaluation.
\newblock {\em arXiv preprint arXiv:1708.00055}.

\bibitem[Chiesi and Primi, 2009]{recency}
Chiesi, F. and Primi, C. (2009).
\newblock Recency effects in primary-age children and college students.
\newblock {\em International Electronic Journal of Mathematics Education},
  4(3):259--279.

\bibitem[Devlin et~al., 2018]{bert}
Devlin, J., Chang, M.-W., Lee, K., and Toutanova, K. (2018).
\newblock Bert: Pre-training of deep bidirectional transformers for language
  understanding.
\newblock {\em arXiv preprint arXiv:1810.04805}.

\bibitem[Doygun and Gulec, 2012]{category2}
Doygun, O. and Gulec, S. (2012).
\newblock The problems faced by university students and proposals for solution.
\newblock {\em Procedia-Social and Behavioral Sciences}, 47:1115--1123.

\bibitem[Gilbert and Hutto, 2014]{vader}
Gilbert, C. and Hutto, E. (2014).
\newblock Vader: A parsimonious rule-based model for sentiment analysis of
  social media text.
\newblock In {\em Eighth International Conference on Weblogs and Social Media
  (ICWSM-14). Available at (20/04/16) http://comp. social. gatech.
  edu/papers/icwsm14. vader. hutto. pdf}, volume~81, page~82.

\bibitem[Head and Eisenberg, 2011]{alison}
Head, A. and Eisenberg, M. (2011).
\newblock How college students use the web to conduct everyday life research.
\newblock {\em First Monday}, 16(4).

\bibitem[Hochreiter and Schmidhuber, 1997]{lstm}
Hochreiter, S. and Schmidhuber, J. (1997).
\newblock Long short-term memory.
\newblock {\em Neural computation}, 9(8):1735--1780.

\bibitem[Jelodar et~al., 2020]{jelodar}
Jelodar, H., Wang, Y., Orji, R., and Huang, H. (2020).
\newblock Deep sentiment classification and topic discovery on novel
  coronavirus or covid-19 online discussions: Nlp using lstm recurrent neural
  network approach.
\newblock {\em arXiv preprint arXiv:2004.11695}.

\bibitem[Kabir and Madria, 2020]{kabir2020coronavis}
Kabir, M. and Madria, S. (2020).
\newblock Coronavis: A real-time covid-19 tweets analyzer.
\newblock {\em arXiv preprint arXiv:2004.13932}.

\bibitem[Kushin and Yamamoto, 2010]{politics2}
Kushin, M.~J. and Yamamoto, M. (2010).
\newblock Did social media really matter? college students' use of online media
  and political decision making in the 2008 election.
\newblock {\em Mass Communication and Society}, 13(5):608--630.

\bibitem[Lai et~al., 2017]{race}
Lai, G., Xie, Q., Liu, H., Yang, Y., and Hovy, E. (2017).
\newblock Race: Large-scale reading comprehension dataset from examinations.
\newblock {\em arXiv preprint arXiv:1704.04683}.

\bibitem[Li and Han, 2013]{cosine}
Li, B. and Han, L. (2013).
\newblock Distance weighted cosine similarity measure for text classification.
\newblock In {\em International Conference on Intelligent Data Engineering and
  Automated Learning}, pages 611--618. Springer.

\bibitem[Liu et~al., 2019]{roberta}
Liu, Y., Ott, M., Goyal, N., Du, J., Joshi, M., Chen, D., Levy, O., Lewis, M.,
  Zettlemoyer, L., and Stoyanov, V. (2019).
\newblock Roberta: A robustly optimized bert pretraining approach.
\newblock {\em arXiv preprint arXiv:1907.11692}.

\bibitem[Loper and Bird, 2002]{nltk}
Loper, E. and Bird, S. (2002).
\newblock Nltk: the natural language toolkit.
\newblock {\em arXiv preprint cs/0205028}.

\bibitem[Loria, 2018]{textblob}
Loria, S. (2018).
\newblock textblob documentation.
\newblock {\em Release 0.15}, 2.

\bibitem[Oyebode et~al., 2020]{oyebode}
Oyebode, O., Ndulue, C., Adib, A., Mulchandani, D., Suruliraj, B., Orji, F.~A.,
  Chambers, C., Meier, S., and Orji, R. (2020).
\newblock Health, psychosocial, and social issues emanating from covid-19
  pandemic based on social media comments using natural language processing.
\newblock {\em arXiv preprint arXiv:2007.12144}.

\bibitem[Paul et~al., 2012]{academic}
Paul, J.~A., Baker, H.~M., and Cochran, J.~D. (2012).
\newblock Effect of online social networking on student academic performance.
\newblock {\em Computers in Human Behavior}, 28(6):2117--2127.

\bibitem[Reimers and Gurevych, 2019]{sen-bert}
Reimers, N. and Gurevych, I. (2019).
\newblock Sentence-bert: Sentence embeddings using siamese bert-networks.
\newblock {\em arXiv preprint arXiv:1908.10084}.

\bibitem[Schroff et~al., 2015]{schroff}
Schroff, F., Kalenichenko, D., and Philbin, J. (2015).
\newblock Facenet: A unified embedding for face recognition and clustering.
\newblock In {\em Proceedings of the IEEE conference on computer vision and
  pattern recognition}, pages 815--823.

\bibitem[Sin et~al., 2019]{relation}
Sin, M., Pyeon, H., Kim, H., and Moon, J. (2019).
\newblock Interpersonal relationship, body image, academic achievement
  according to sns use time of college students.
\newblock {\em The Journal of the Convergence on Culture Technology},
  5(1):257--264.

\bibitem[Thelwall et~al., 2012]{thelwall2}
Thelwall, M., Buckley, K., and Paltoglou, G. (2012).
\newblock Sentiment strength detection for the social web.
\newblock {\em Journal of the American Society for Information Science and
  Technology}, 63(1):163--173.

\bibitem[Tsakalidis et~al., 2018]{mental2}
Tsakalidis, A., Liakata, M., Damoulas, T., and Cristea, A.~I. (2018).
\newblock Can we assess mental health through social media and smart devices?
  addressing bias in methodology and evaluation.
\newblock In {\em Joint European Conference on Machine Learning and Knowledge
  Discovery in Databases}, pages 407--423. Springer.

\bibitem[Vaswani et~al., 2017]{transformers}
Vaswani, A., Shazeer, N., Parmar, N., Uszkoreit, J., Jones, L., Gomez, A.~N.,
  Kaiser, {\L}., and Polosukhin, I. (2017).
\newblock Attention is all you need.
\newblock In {\em Advances in neural information processing systems}, pages
  5998--6008.

\bibitem[Wang et~al., 2018]{glue}
Wang, A., Singh, A., Michael, J., Hill, F., Levy, O., and Bowman, S.~R. (2018).
\newblock Glue: A multi-task benchmark and analysis platform for natural
  language understanding.
\newblock {\em arXiv preprint arXiv:1804.07461}.

\bibitem[Williams et~al., 2017]{mnli}
Williams, A., Nangia, N., and Bowman, S.~R. (2017).
\newblock A broad-coverage challenge corpus for sentence understanding through
  inference.
\newblock {\em arXiv preprint arXiv:1704.05426}.

\bibitem[Yang and Lee, 2020]{politics1}
Yang, C.-c. and Lee, Y. (2020).
\newblock Interactants and activities on facebook, instagram, and twitter:
  Associations between social media use and social adjustment to college.
\newblock {\em Applied Developmental Science}, 24(1):62--78.

\end{thebibliography}

% \section*{\uppercase{Appendix}}

% \noindent If any, the appendix should appear directly after the
% references without numbering, and not on a new page. To do so please use the following command:
% \textit{$\backslash$section*\{APPENDIX\}}

\end{document}